\documentclass[runningheads]{llncs}
\usepackage[T1]{fontenc}
\usepackage{graphicx}
\usepackage{booktabs}
\usepackage{amssymb}
\usepackage{multirow}

\usepackage{todonotes}

\newcommand{\bs}{$\blacksquare$}

\newcommand{\ct}{\texttt{CheckThat!}}

\usepackage{url}
\usepackage{soul}
\usepackage{xcolor} 
\usepackage{todonotes}
\definecolor{blue}{rgb}{0,0, 0.6}
\definecolor{dkgreen}{rgb}{0,0.6,0}
\definecolor{gray}{rgb}{0.5,0.5,0.5}
\definecolor{mauve}{rgb}{0.58,0,0.82}
\definecolor{mauve}{rgb}{0,0,0}
\definecolor{black}{rgb}{0,0,0}
\definecolor{tri}{rgb}{.25,.88,.82}
\definecolor{lilac}{rgb}{0.85,0.64,0.85}
\definecolor{lightblue}{rgb}{0.53, 0.81, 0.98}
\definecolor{lightskyblue}{rgb}{0.53, 0.81, 0.98}

\begin{document}
\title{The CLEF-2026 CheckThat! Lab: Advancing Multilingual Fact-Checking}
%
\titlerunning{The CLEF-2026 CheckThat! Lab}
%
\author{%
Julia Maria Struß\inst{1}\orcidID{0000-0001-9133-4978} \and
Sebastian Schellhammer\inst{2,3}\orcidID{0009-0001-6413-5823} \and
Stefan Dietze\inst{2,3}\orcidID{0009-0001-4364-9243} \and
Venktesh V \inst{4} \orcidID{0000-0001-5885-2175} \and 
Vinay Setty \inst{5} \orcidID{0000-0002-9777-6758} \and 
Tanmoy Chakraborty\inst{6}\orcidID{0000-0002-0210-0369} \and
Preslav Nakov\inst{7}\orcidID{0000-0002-3600-1510} \and
Avishek Anand \inst{8} \orcidID{0000-0002-0163-0739} \and 
Primakov Chungkham \inst{9} \orcidID{0009-0005-8732-5338} \and 
Salim Hafid\inst{10}\orcidID{0000-0002-1775-8542} \and
Dhruv Sahnan\inst{7}\orcidID{0000-0002-5205-8269} \and 
Konstantin Todorov\inst{11}\orcidID{0000-0002-9116-6692}
}
\authorrunning{J.\,M. Struß et al.}
%
\institute{%
University of Applied Sciences Potsdam, Germany \and
GESIS - Leibniz Institute for the Social Sciences, Cologne, Germany \and
Heinrich-Heine-University D\"usseldorf, Germany  \and
Stockholm University, Sweden \and 
University of Stavanger, Norway \and 
Indian Institute of Technology Delhi, India \and 
Mohamed bin Zayed University of Artificial Intelligence, Abu Dhabi \and 
Delft University of Technology, The Netherlands \and
Independent Researcher \and
médialab Sciences Po, Paris, France \and
University of Montpellier, LIRMM, CNRS, Montpellier, France\\
\email{https://checkthat.gitlab.io}
}

\maketitle              
\begin{abstract}
The \ct~lab aims to advance the development of innovative technologies combating disinformation and manipulation efforts in online communication across a multitude of languages and platforms. While in early editions the focus has been on core tasks of the verification pipeline (check-worthiness, evidence retrieval, and verification), in the past three editions, the lab added additional tasks linked to the verification process. In this year's edition, the verification pipeline is at the center again with the following tasks:
Task~1 on source retrieval for scientific web claims (a follow-up of the 2025 edition),
Task~2 on fact-checking numerical and temporal claims, which adds a reasoning component to the 2025 edition, and 
Task~3, which expands the verification pipeline with generation of full-fact-checking articles.
These tasks represent challenging classification and retrieval problems as well as generation challenges at the document and span level, including multilingual settings. 
\keywords{disinformation \and
fact-checking \and
claim source retrieval \and
generating fact-checking articles}
\end{abstract}
%
%
\newpage

\section{Introduction}
\label{sec:intro}
\begin{figure}[b]
\includegraphics[width=\textwidth]{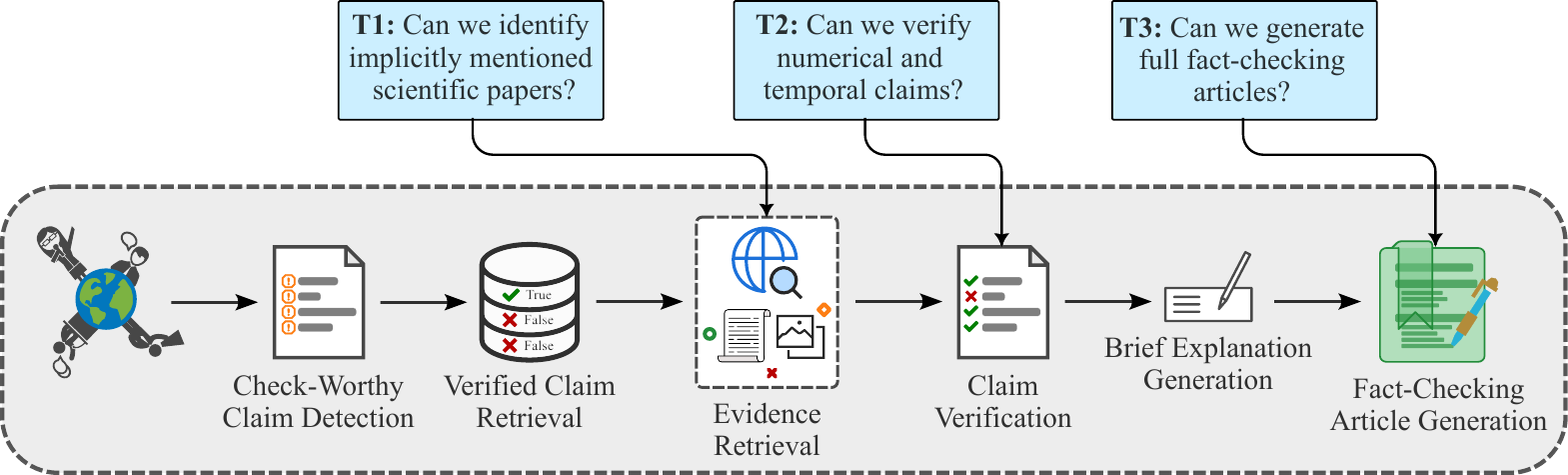}
\caption{Overview of the \ct{} verification pipeline, which features the core fact-checking tasks along with the tasks we address in the 2026 edition of the lab.
}
\label{fig:pipeline}
\end{figure}

The \ct{} lab aims at fostering the development of technology to combat the spreading of mis- and disinformation across multiple platforms and languages in online discourses. Over the past eight iterations
(e.g., see \cite{clef-checkthat:2025-lncs,clef-checkthat:2024-lncs}),
a multitude of challenges and tasks along the main steps of the verification pipeline used by \textit{journalists} and \textit{fact-checkers} (see Fig.~\ref{fig:pipeline}) have been offered: Starting with a given document or a claim, the first step is to evaluate its check-worthiness, i.e.,~whether human effort should be allocated for checking its veracity, which is a task that experts have repeatedly emphasized as being of particular importance to them in the past. As claims that have been proven false before are repeatedly reformulated and circulated, retrieving closely related, previously fact-checked claims is the next step in the verification process. Retrieving additional evidence from diverse sources to verify the claim and finally decide whether the claim is factually true or not, with various degrees in between, follows. Ultimately, the verification process is completed by writing a detailed article that provides the evidence and the final decision, a step which has been added to the \ct{} verification pipeline for the first time in the 2026 edition of the lab.

The three tasks proposed for the 2026 edition cover three of the main tasks in the pipeline, highlighted in Figure \ref{fig:pipeline}, and span a total of five different languages (see in Table~\ref{tab:lang}):



\noindent
\textbf{Task~1 Source Retrieval for Scientific Web Claims,} is related to the second and third task of the verification pipeline and aims to detect different types of references or mentions to scientific work as well as identifying the original study a social media post refers to 
(cf.~Section~\ref{sec:task1}).

\noindent
\textbf{Task~2 Fact-Checking Numerical and Temporal Claims} addresses the claim verification task of the pipeline, focusing on numerical claims (cf.~Section~\ref{sec:task2}).

\noindent
\textbf{Task~3 Generating Full-Fact-Checking Articles:} introduces a new, final task into the CLEF CheckThat! lab pipeline, which attempts to automate the fact-checking article writing process (cf.~Section~\ref{sec:task3}).

\begin{table}
\centering
\caption{Languages targeted in the three tasks of the \ct{} 2026 edition. 
}\label{tab:lang}
\begin{tabular}{@{}lccccc@{}}
    \toprule
        Task	&	Arabic	  &	English	  &	German	  &	French	&	Spanish \\\midrule
        Task 1	&		      &	\bs	      &	\bs	      &	\bs	    &	        \\
        Task 2	&	\bs	      &	\bs	      &		      &		    &	\bs     \\
        Task 3	&		      &	\bs	      &		      &		    &	        \\
    \bottomrule
    \end{tabular}
\end{table}

\section{Task 1: Source Retrieval for Scientific Web Claims}
\label{sec:task1}
\paragraph{\textbf{Motivation}}

Scientific web discourse, i.e., discourse about scientific claims or resources on the social web, has increased substantially throughout the past years and has been studied across a wide range of disciplines \cite{bruggemann2020post,dunwoody2021science}.
However, the scientific web discourse is usually informal, including posts such as \textit{``covid vaccines just don't work on children''}. The scientific studies at the center of debates from which claims originate are usually not cited explicitly but only mentioned informally, such as \textit{``Stanford study shows that vaccines don't work''}. This poses challenges both from a computational perspective when mining social media or computing altmetrics, but also from a societal perspective, leading to poorly informed online debates \cite{rocha2021impact}. For fact-checking, identifying the source of a claim is important, especially for false claims where counter-evidence may not exist, because human fact-checkers use the source to verify a claim by assessing the quality of the source publication and checking whether the claim accurately reflects its findings \cite{glockner2022missing}. Based on this motivation, we propose the task of Source Retrieval for Scientific Web Claims:

\paragraph{\textbf{Task definition}}
 \emph{Given a social media post that contains a scientific claim and an implicit reference to a scientific paper (mentions it without a URL), retrieve the mentioned paper from a pool of candidate papers. (Languages: English, German, and French.)}

\paragraph{\textbf{Data}}
Ground-truth datasets for training and evaluation will be provided in three languages: English, German, and French. Each dataset consists of pairs of social media posts from X and the scientific studies they implicitly reference. For English, we reuse the existing 15,699 data pairs from \cite{clef-checkthat:2025:task4} for training, while new data will be annotated for the evaluation set. For German and French, new data will be annotated for both training and evaluation sets, with approximately 1,500 pairs per language. For each language, a query dataset will be provided, containing social media posts, along with a single collection set that includes information about the scientific studies, which will be used across all three languages.

\paragraph{\textbf{Evaluation}}
For each language, the participating systems will be evaluated in terms of Mean Reciprocal Rank at 5 (\textit{MRR@5}), which is an established measure for evaluating known-item search tasks. It focuses on ranking the correct study within the first five positions, as users are only rarely examine more results.

\section{Task 2: Fact-Checking Numerical and Temporal Claims}
\label{sec:task2}


 \paragraph{\textbf{Motivation}}

In recent years, tremendous progress has been made in automated fact-checking approaches~\cite{guo2022survey,setty2024factcheck}, along with the development of large-scale benchmarks~\cite{multifc,schlichtkrull2023averitec}, for tackling misinformation at scale. Automating fact-checking is inherently knowledge-intensive, as it requires reasoning over heterogeneous evidence sources to arrive at a reliable verdict. This challenge is further amplified for claims involving numerical information \cite{venktesh2024quantemp}, which require numerical contextualization and precise quantitative reasoning. Moreover, numerical claims are prone to the \emph{Numeric-Truth Effect} \cite{sagara2009consumer}, whereby the presence of numbers induces a false sense of credibility, thereby accelerating the spread of misinformation. Although Large Language Models (LLMs) have achieved remarkable advances on knowledge-intensive tasks, recent studies demonstrate that verifying numerical claims remains challenging for LLMs than verifying non-numerical ones~\cite{venktesh2024quantemp,Aly:2021:NeurIPs}. Motivated by recent advances in test-time reasoning \cite{chungkham-etal-2025-think}, we propose a test-time scaling framework with the goal of training more effective verification models to enhance the performance of LLMs on fact-checking numerical claims.

\paragraph{\textbf{Task definition}} \emph{This task performs test-time scaling for fact-verification. Given a claim, relevant evidences and reasoning traces with corresponding verdicts, the task is to rank the reasoning traces based on their utility in leading to the correct verdict and also output the final verdict from the top-ranked reasoning traces. (Languages: English, Spanish and Arabic.)}

The task focuses on verifying naturally occurring claims that contain quantitative quantities and/or temporal expressions by improving the reasoning capabilities of LLMs through test-time scaling. Unlike previous editions, which primarily focused only on fact-checking accuracy, this year's task explicitly incorporates rationale generation  into the evaluation. This is achieved in the form of re-ranking reasoning traces from the LLM, assessing both the correctness of the predicted veracity and the quality of the underlying reasoning traces.

While the claims are drawn from the previous iteration of the task, the overall setup differs substantially. Specifically, we introduce a test-time scaling framework designed to enhance LLM reasoning for claim verification. Given a claim and its associated evidence, multiple reasoning traces are generated using an LLM with varying temperature values to induce diversity. Redundant traces are subsequently removed through a de-duplication step. Based on this data, partici-pants are required to train a verifier model that ranks the reasoning traces for each test claim and derives a final verdict from the top-ranked traces. To avoid leakage and ensure rigorous evaluation, we will release new test sets.


\begin{table}[t]
\footnotesize\centering
\caption{Dataset statistics for task 2.}
\label{tab:my_label}
\begin{tabular}{l@{ }@{ }@{ }r}
\toprule
\bf Language   & \bf \# of claims    \\\midrule
English &  8,000 \\
Spanish     &  2,808                       \\
Arabic     &    3,260                   \\
\bottomrule
\end{tabular}
\end{table}

\paragraph{\textbf{Data}}
We collect the dataset from various fact-checking websites, completed with detailed metadata and an evidence corpus sourced from the web.  We use the English dataset released in \cite{venktesh2024quantemp}, as well as the 3,260  Arabic and 2,808 Spanish claims from CLEF 2025 \ct~\cite{clef-checkthat:2025-lncs} with corresponding evidence. Table~\ref{tab:my_label} summarizes the statistics of the dataset.\\

Each claim in the training and the validation datasets comprises 20 reasoning traces generated by \textbf{gpt-4o-mini} that can be used by participants to train a verification model, for selecting the best reasoning traces that lead to the correct verdict. Apart from top-10 evidences, the entire evidence corpus will also be provided to the participants so they can run their own retrieve and re-rank pipelines for providing additional context to the verification model.
Each claim is annotated with relevant reasoning traces that lead to the correct verdict (ground truth). The  traces re-ranked by the verifier can be evaluated based on their ability to identify such relevant traces.

\paragraph{\textbf{Evaluation}} We use Recall@k and MRR@k to measure the quality of the reasoning traces ranked by the verifier model, with $k=5$.
For claim verification, we use macro-averaged F1 and classwise F1 scores. 
Participants will be judged based on high performance on both macro F1 and Recall@$k$. The final score is obtained by averaging ranking across these measures (macro F1 and Recall@k, with $k=5$).

\section{Task 3: Generating Full-Fact-Checking Articles}
\label{sec:task3}


\paragraph{\textbf{Motivation}} Previous work on automated fact-checking has focused on five main tasks: 
(\emph{i})~check-worthy claim detection,
(\emph{ii})~previously verified claim retrieval,
(\emph{iii})~evidence retrieval,
(\emph{iv})~claim verification, and
(\emph{v})~justification generation~\cite{vlachos-riedel-2014-fact,guo2022survey,assist-fc-survey-preslav}.
However, professional fact-checkers also undertake the critical task of \textit{publishing detailed fact-checking articles}, which guide readers toward a clear understanding of the claim by presenting factual arguments and explaining how they lead to the final verdict~\cite{graves-anatomyofFC-2017}.
Writing fact-checking articles is time-consuming, as it involves synthesizing information from various sources, penning it down in an easy to understand draft, and refining it multiple times to ensure quality and correctness.
Recently, we proposed the generation of fact-checking articles automatically using large language models~\cite{sahnan2025llmsautomatefactcheckingarticle}.
While LLMs can generate fluent, long-form text ~\cite{wang2024generating,xie-etal-2023-next,yang-etal-2023-doc}, our evaluations show that LLM-generated fact-checking articles considerably lag behind expert-written articles, uncovering several issues, but especially their inability to construct coherent, evidence-based arguments that properly explain the claim's veracity.

\paragraph{\textbf{Task definition}} \emph{Given a claim, its veracity, and a set of evidence documents consulted for fact-checking the claim, generate a \textit{full fact-checking article with inline citations}. (Language: English.)}


Note that, we do not address automatic evidence retrieval or claim verification here: the claim, its veracity, and the evidence set to use are all explicitly provided.

\paragraph{\textbf{Data}} We reuse the WatClaimCheck dataset~\cite{khan-etal-2022-watclaimcheck} for training and validation, and the union of the already publicly available ExClaim~\cite{zeng-gao-2024-justilm} dataset and currently private AmbiguousSnopes~\cite{sahnan2025llmsautomatefactcheckingarticle} dataset for testing.
Overall, the training set comprises 26k data examples, and the validation set comprises 3.3k data examples.
The test set contains 1.2k data examples in total, comprising 987 from ExClaim and 220 from AmbiguousSnopes.
Each example corresponds to a claim and comprises the claim's veracity label, a set of evidence documents as listed by the fact-checking expert on the source website, and the expert-written fact-checking article, which we treat as the ground truth.


\paragraph{\textbf{Evaluation}} As a primary evaluation measure, we use the mean of the following evaluation measures to assess the quality of the generated fact-checking article:
(\emph{i})~\emph{entailment score}, which is a reference-based metric that measures if the generated text is entailed by the reference~\cite{sahnan2025llmsautomatefactcheckingarticle},
(\emph{ii})~\emph{citation correctness}, which verifies if a text attributed to a citation can be entailed by the corresponding evidence,
and (\emph{iii})~\emph{citation completeness}, which computes the proportion of input evidence that is correctly cited in the generated text.
As a secondary measure, we also assess writing quality by computing Elo ratings for the participants at the end of the testing phase, through a setup similar to Chatbot Arena~\cite{chiang2024chatbotarenaopenplatform}, but using LLM-as-a-judge~\cite{llm_as_a_judge-2023,kim2024prometheus} in our case.
We choose to use reference-free evaluation measures to assess the utility of the participating systems for professional fact-checkers, rather than how similar the generated articles are to the expert-written articles.
This design choice also aims to prevent cheating on the publicly-available subset of our test dataset.

\section{Related Work}
The tasks offered in the different editions of the \ct{} lab are closely related to several tasks at other evaluation initiatives such as SemEval, e.\,g. determining rumor veracity~\cite{derczynski2017semeval,gorrell2019semeval}, stance detection~\cite{mohammad2016semeval}, fact-checking in community question-answering forums~\cite{mihaylova2019semeval}, propaganda detection in various contexts such as English news articles~\cite{da2020semeval}, English memes~\cite{dimitrov2024semeval,dimitrov2021semeval2021}, and Arabic news paragraphs, social media posts, and memes~\cite{hasanain-etal-2023-araieval,hasanain2024araieval}, as well as multi- and cross-lingual fact-checked claim retrieval~\cite{peng-etal-2025-semeval-claim-retrieval}, emphasizing the focus on multi-linguality, also including low-resourced languages.
The tasks offered in the \ct{} lab are further related to the FEVER tasks~\cite{schlichtkrull-etal-2024-automated,thorne2018fever} on fact extraction and verification, the Fake News Challenge~\cite{hanselowski-etal-2018-retrospective,malliga2023overview,FNC1}, and the detection of online hostile posts~\cite{patwa2021overview}. Furthermore, it aligns with the FakeNews task at MediaEval~\cite{pogorelov2020fakenews,pogorelov2022fakenews_mediaeval}.
Other recent shared tasks include claim detection and span identification in social media posts~\cite{sundriyal2023overview} as well as multi-modal fake news detection~\cite{suryavardan2022findings}, further emphasizing the growing focus on analyzing and verifying information across diverse modalities and platforms.

\section{Conclusion and Future Work}
\label{sec:conclusions}

We presented the tasks of the 2026 edition of the \ct{} lab, featuring complementary tasks to assist journalists and fact-checkers along the fact-checking pipeline: from retrieving informally mentioned sources of scientific web claims to verifying numerical and temporal claims and ultimately generating full-fact-checking articles - a task that has been newly added to the \ct{} verification pipeline. In line with the CLEF mission, two of the three tasks are multilingual by offering tasks in five languages, namely Arabic, English, German, French, and Spanish.

In future work, we plan to expand the multilingual coverage, particularly for low-resource languages. We also plan to explore stronger cross-document and numerical reasoning, improved handling of implicit evidence, and closer alignment of task design and evaluation with real-world fact-checking workflows to maximize practical impact.

\section*{Acknowledgments}
\begin{footnotesize}
The work of Stefan Dietze, Konstantin Todorov, Salim Hafid, and Sebastian Schellhammer is partially funded under the AI4Sci grant, co-funded by MESRI (France, grant UM-211745), BMBF (Germany, grant 01IS21086), and the French National Research Agency (ANR).
The work of Dhruv Sahnan, Tanmoy Chakraborty, and Preslav Nakov is supported by the Multi-Institutional Faculty Interdisciplinary Research Project\linebreak (MFIRP) between IIT Delhi and MBZUAI.
The responsibility for the contents of this publication lies with the authors.

\end{footnotesize}

\section*{Disclosure of Interests}
The authors have no competing interests to declare that are relevant to the content of this article.
%
%
%
\bibliographystyle{splncs04}
\bibliography{clef20_checkthat,clef19_checkthat,clef18_checkthat,clef21_checkthat,clef22_checkthat,clef23_checkthat,custom,sigproc,clef24_checkthat,clef25_checkthat,related_shared_tasks}
\end{document}